# Enhancing Steganographic Text Extraction: Evaluating the Impact of NLP Models on Accuracy and Semantic Coherence

Mingyang Li[a], Maoqin Yuan*[a], Luyao Li[b], Han Pengsihua[b]

[a] School of Liberal Arts and Sciences, China University of Petroleum-Beijing at Karamay, Karamay 834000,Xinjiang, China;

[b] School of Business Administration, China University of Petroleum-Beijing at Karamay, Karamay 834000,Xinjiang, China.

* Corresponding author: mqyuan@cupk.edu.cn

## ABSTRACT

*This study discusses a new method combining image steganography technology with Natural Language Processing (NLP) large models, aimed at improving the accuracy and robustness of extracting steganographic text. Traditional Least Significant Bit (LSB) steganography techniques face challenges in accuracy and robustness of information extraction when dealing with complex character encoding, such as Chinese characters. To address this issue, this study proposes an innovative LSB-NLP hybrid framework. This framework integrates the advanced capabilities of NLP large models, such as error detection, correction, and semantic consistency analysis, as well as information reconstruction techniques, thereby significantly enhancing the robustness of steganographic text extraction. Experimental results show that the LSB-NLP hybrid framework excels in improving the extraction accuracy of steganographic text, especially in handling Chinese characters. The findings of this study not only confirm the effectiveness of combining image steganography technology and NLP large models but also propose new ideas for research and application in the field of information hiding. The successful implementation of this interdisciplinary approach demonstrates the great potential of integrating image steganography technology with natural language processing technology in solving complex information processing problems.*

## 1. INTRODUCTION

In the fields of digital communication and data security, image steganography plays an indispensable role [1][2]. As a means of hiding information, it allows for the transmission of data without drawing attention, complementing encryption technology. With the growing demand for information security in recent years, research and applications of image steganography have attracted widespread attention [3][4].

Among various steganography methods, the Least Significant Bit (LSB) algorithm has become a hot research topic due to its simplicity and efficiency [5][6]. This method embeds secret information by modifying the least significant bit of image pixels, usually causing minimal visual impact on the image [7]. However, when steganographic images undergo common image processing such as compression, scaling, and noise addition, the embedded information often gets damaged, affecting the integrity and readability of the information [8].

Despite the excellent performance of LSB steganography in hiding information, it has evident shortcomings in the robustness of information extraction [9]. This issue becomes particularly pronounced when dealing with Chinese characters, whose complex encoding can make it difficult for traditional error recovery techniques to accurately reconstruct the original information once it is damaged [10]. Therefore, seeking a method that improves the robustness of steganographic text extraction has become particularly important. The rapid development of Natural Language Processing (NLP) technology in recent years, especially the emergence of large pretrained models like GPT-3.5 and BERT, offers a new perspective for addressing this issue [11][12]. These models show remarkable abilities in semantic understanding, text generation, error detection, and correction, bringing new possibilities for improving the accuracy and robustness of steganographic text extraction.

This study aims to explore the potential of combining LSB steganography technology with large NLP models, particularly in the extraction process of Chinese character steganographic text. We propose a new method (LSB-NLP hybrid framework) that utilizes large NLP models for error detection and correction, semantic consistency analysis, and information reconstruction to address the damage issues in the steganography process. This paper will detail the design, implementation, and evaluation of this method, as well as a comparative analysis with traditional methods, to demonstrate its advantages and potential in improving the accuracy of steganographic text extraction.

## 2. CONSTRUCTION OF A BENCHMARK MODEL FOR LSB CHINESE STEGANOGRAPHY

In this section, we delve into the establishment of an efficient and covert method of information hiding, namely using the Least Significant Bit (LSB) steganography technique to embed Chinese text into images.

### 2.1 Binary Conversion of Text

In the steganography process, the primary task is converting Chinese text into a binary string suitable for embedding in an image. Given the relative complexity of Chinese character encoding, we adopt UTF-8 encoding for this conversion, ensuring that each Chinese character is accurately transformed into its corresponding binary sequence. The specific conversion steps are as follows:

1.UTF-8 Encoding Conversion: Firstly, the Chinese text is converted using UTF-8 encoding. This step ensures that all characters in the text can be represented in bytes.

2.Byte Sequence Processing: Each UTF-8 encoded character is converted into one or more bytes. We extract these byte sequences for further processing.

3.Byte to Binary Conversion: Next, each byte is further converted into an 8-bit binary representation.

This conversion process not only ensures that Chinese characters can be correctly encoded but also provides a solid foundation for the subsequent embedding process. The use of UTF-8 encoding allows us to handle various Chinese characters and special symbols, thereby enhancing the applicability and flexibility of the steganography system.

### 2.2 Steganography of Text

We embed the binary data processed in 2.1 into the pixels of the target image. The primary characteristics of the UTF-8-LSB algorithm are as follows:

1.Pixel-level Embedding: In this study, images are processed pixel by pixel, starting from the top left corner and scanning row by row to the bottom right. For each pixel, we slightly adjust the least significant bit of its RGB value to embed the binary data. This minor adjustment barely affects the visual quality of the image. By making only small changes at the least significant bit, we maintain the image's appearance virtually unchanged.

2.Data Embedding Algorithm: To achieve efficient data embedding, we have designed the UTF-8-LSB algorithm. These algorithms ensure that a significant amount of data can be successfully embedded even in localized areas of the image, thus improving the efficiency of steganography. Our algorithms take into account the characteristics of the image to minimize the impact on image quality.

3.Preservation of Visual Quality: During the steganography process, special attention is paid to preserving the visual quality of the image. Our algorithms and methods are designed to ensure that the steganographic image is nearly indistinguishable from the original, preventing the detection of embedded data. This is due to the nature of LSB embedding in the least significant bit, which makes the human eye insensitive to the changes before and after.

### 2.3 Extraction of Text Post-Steganography

In steganography, the extraction of information is one of the key components of steganographic techniques. This section will detail how to successfully extract the embedded Chinese text from a steganographic image. The specific extraction steps are as follows:

1.Image Data Reading: The first step in the extraction process is to read the image containing the steganographic information. At this stage, we only obtain the pixel data of the image without altering its visual characteristics.

2.Pixel-level Extraction: Starting from the top left corner of the image and scanning row by row to the bottom right, we extract the least significant bit of the RGB value of each pixel. This method allows for the accurate restoration of the embedded binary data.

3.Data Decoding: The extracted binary data needs to be further decoded to restore the Chinese text. We use the same encoding format as during embedding, i.e., UTF-8 encoding, to convert the binary data back into Chinese characters.

4.Integrity Check of Information: During the extraction process, we perform an integrity check of the information to ensure that the extracted text is complete. This includes checking for the presence of a predefined end marker to signify the successful extraction of all embedded text.

## 2.4 Accuracy Test of Extraction

In the field of information steganography, extraction accuracy is one of the key indicators of the performance of a steganography system. Before testing, we perform the extraction process of the post-steganography image, restoring the steganographic information to text. The extracted text is then compared with the original Chinese text. We will use a comparison algorithm to compare the extracted text with the original text character by character to determine their consistency. The extraction accuracy can be calculated using the following formula:

$$Accuracy = N/T$$

Here, "N" represents the number of characters correctly extracted, and "T" represents the total number of characters.

## 2.5 Model Results

In this study, we used the standard test image from Lenna [13] for steganography testing. To evaluate the effectiveness and covertness of the steganography technique, a text of 10,000 characters in length was embedded into this image. Visually, the steganographic image is almost indistinguishable from the original, demonstrating that the steganography technique successfully hides a substantial amount of textual information while maintaining image quality.

As part of the quantitative evaluation, Mean Square Error (MSE) was used as the primary metric. The calculated MSE value is 0.2268, which is considered extremely low in the field of steganography, indicating very minimal pixel-level differences between the original and the steganographic images. This low MSE value further attests to the high covertness of the steganography technique used and its minimal impact on the original image quality.

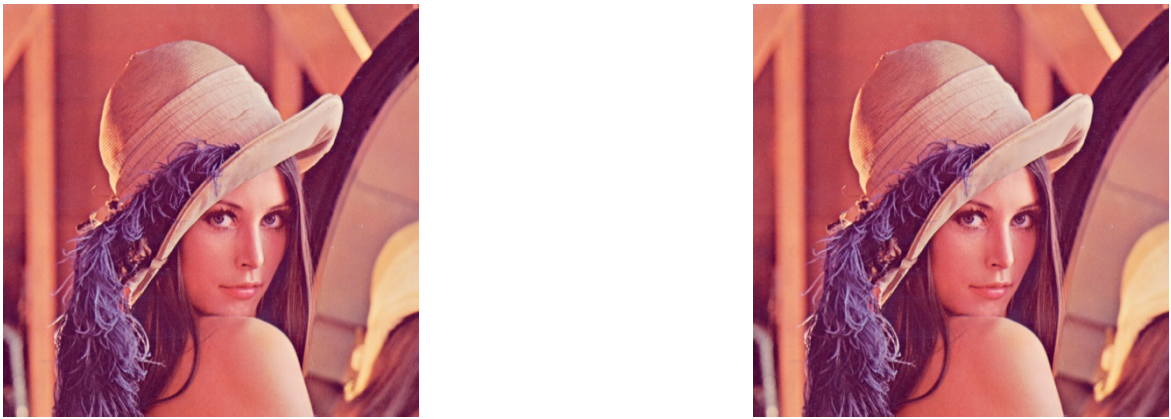

Figure 1. Original Image (left) and Image with 10,000 Characters Embedded (right)

The accuracy of text extraction from the steganographic image reached 100%, demonstrating the efficiency and accuracy of the steganography system in information extraction. This implies that the system is capable of losslessly recovering hidden information, ensuring data integrity and reliability in transmission.

In summary, these results indicate that the steganography model excels in covertness, preservation of image quality, and accuracy of data recovery. Therefore, it can be considered efficient and practical for high-capacity data steganography tasks.

## 3. CONSTRUCTION OF AN NLP MODEL FOR STEGANOGRAPHY EXTRACTION

In this section, we introduce the construction and application of a Natural Language Processing (NLP) model specifically designed for extracting steganographic information. This part will explore the core concepts of the NLP model, its improvements in handling garbled text extraction, and testing of the model.

### 3.1 Introduction to the NLP Model

We have chosen an NLP model based on the Transformer architecture to address the challenges of extracting steganographic information. Known for its superior sequential modeling capabilities and parallel computing advantages, the Transformer model is suitable for various NLP tasks. The architecture of this NLP model is based on the Self-Attention mechanism, allowing the model to maintain global context in processing variable-length input sequences. The model captures semantic information in the input sequence through multiple layers of self-attention and feed-forward neural networks, mapping it to a high-dimensional embedding space. The input structure of the model includes garbled text extracted from steganographic images, which are divided into different tokens for the model to process. The output structure involves understanding the garbled text and mapping it back to natural language text.

During training, we used a large-scale Chinese text dataset to fine-tune the model's parameters to adapt to the task of extracting steganographic information. The training process involved optimizing the model's loss function to maximize its understanding and accuracy in extracting steganographic information. Unlike traditional NLP, our model includes a fine-tuning mechanism and uses a large amount of correct steganographic answers and data extracted via LSB for testing, significantly improving the model's accuracy in extracting image steganography. This is due to its self-attention mechanism, which enables it to handle the contextual relationships in garbled text, thereby improving accuracy. Additionally, the model's deep structure allows it to learn complex semantic representations, aiding in handling semantic consistency in steganographic information. The improvement in accuracy is attributed to the low correlation between the garbled text and the original text. The NLP model increases the readability of the original text by correcting words based on context. Even if the corrected characters do not match the correct characters, it does not reduce the extraction accuracy.

### 3.2 NLP Changes in Garbled Text Extraction Training

Traditional NLP models may face challenges due to the discontinuity and garbled nature of text resulting from steganography. Therefore, we implemented a series of improvements to enhance the NLP model's understanding and accuracy in extracting garbled text.

In training the NLP model, we improved data preprocessing to adapt to the specific nature of garbled text. Firstly, we introduced a special data cleaning step for handling special characters, symbols, and formats in steganographic information. This helps reduce misunderstandings by the model during extraction. Secondly, we tokenized the garbled text, breaking it down into smaller units corresponding to partial vocabulary, character fragments, or specific encoding sequences. Tokenization allows the model to better capture local patterns and associations in steganographic information, thus improving extraction precision.

To adapt the NLP model to the task of extracting garbled text, we adopted a carefully designed fine-tuning strategy. During the model fine-tuning stage, we introduced labels for steganographic information to guide the model in understanding garbled text. These labels indicate to the model which parts are steganographic information and which are normal text, enabling the model to more accurately identify and extract steganographic information. Furthermore, we conducted multiple rounds of iterative fine-tuning to gradually improve the model's performance on the garbled text extraction task. Each round of fine-tuning involved minor adjustments to the model's parameters to better adapt to the characteristics of steganographic information.

Finally, to further enhance the NLP model's performance, we conducted specific feature engineering for garbled text. This included introducing specially designed features to capture patterns and semantic associations in steganographic information. The goal of these feature engineering efforts was to enhance the model's robustness and accuracy in the task of extracting garbled text.

## 4. INTEGRATION OF LSB CHINESE WITH NLP MODEL

In this section, we combine LSB Chinese steganography with the NLP model to construct a powerful framework for extracting steganographic information. This framework combines the high covertness of LSB steganography with the text comprehension ability of the NLP model, providing a comprehensive and effective solution for the extraction of steganographic information.

## 4.1 LSB-NLP Framework

The core idea of the LSB-NLP framework is to use LSB steganography for embedding Chinese text into the least significant bits of pixels in an image, while utilizing the NLP model for text extraction and restoration. This framework includes the following key steps:

1.Text Embedding (LSB Steganography): First, the Chinese text, after specific encoding and conversion, is embedded into the least significant bit of the pixels in the target image. This step ensures the concealment of the data while not significantly affecting the visual quality of the image.

2.Image Storage: The image with embedded text is saved and can be used for subsequent extraction processes.

3.Text Extraction (NLP Model): In the extraction phase, we use the NLP model to extract the steganographic Chinese text from the embedded image. The NLP model is specially trained to adapt to the garbled and noisy characteristics of steganographic information, thereby improving extraction accuracy.

## 4.2 Testing the Model

This section will detail the testing process and results of the LSB-NLP framework. We will use a series of steganographic images with different characteristics and lengths to evaluate the framework's performance. The primary objective of the test is to verify the accuracy of the framework.

In the tests, we will assess the accuracy and stability of the LSB-NLP framework in extracting different types of steganographic information. We will also examine the framework's performance in the face of different character lengths and noise, to determine its usability in complex scenarios. Under the interference of noise, by embedding texts of different lengths, we compare the accuracy of the LSB steganography with the steganography algorithm based on the LSB-NLP framework as shown in the following graph.

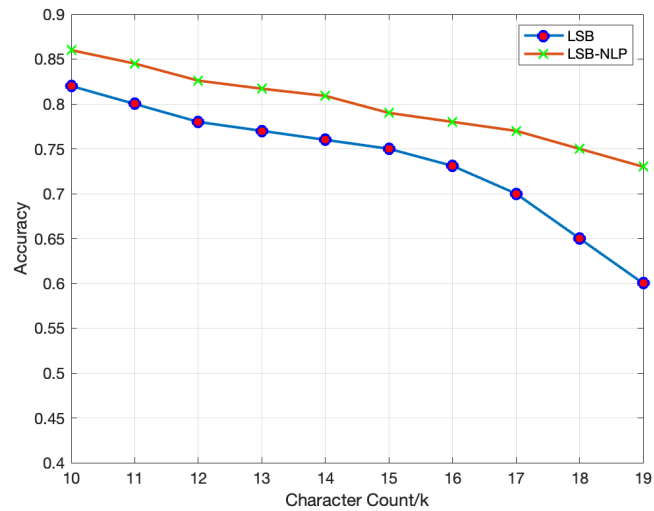

Figure 2. Comparative Analysis of LSB-NLP Framework Versus LSB Benchmark in Steganographic Accuracy

The test results show that the LSB-NLP model outperforms the LSB benchmark model in terms of accuracy. The model achieves high accuracy with different numbers of characters (10k characters - 19k characters). This indicates that the LSB-NLP model can effectively extract steganographic information and has high accuracy.

## 5. CONCLUSIONS

This study aimed to explore the issues of steganography and extraction of Chinese text, proposing an innovative LSB-NLP framework by combining Least Significant Bit (LSB) steganography with Natural Language Processing (NLP) models. Through detailed experiments and performance evaluations, the study reached the following conclusions:

Improved Accuracy and Robustness: By fine-tuning and training the NLP model, this study enhanced the understanding and accuracy of extracting garbled text. Experimental results show that the LSB-NLP framework excels in various steganographic information extraction tasks, particularly demonstrating exceptional robustness in handling garbled and noisy data.

Superiority Over Traditional Methods: Compared to traditional LSB techniques and standalone NLP models, the LSB-NLP framework showed a clear advantage in extraction accuracy. This provides strong support for research and applications in the field of information hiding.

Although significant achievements were made, there are many future research directions worth exploring. The framework's performance can be further optimized, extended to other languages and application areas, and research on robustness and security can be strengthened. Additionally, considering the integration of more advanced deep learning technologies to further enhance the processing capability of steganographic information is a viable avenue.

Overall, this study offers an effective solution to the issues of steganography and extraction of Chinese text, opening new possibilities for research and applications in the field of information hiding. We look forward to future research continuing to advance this field, contributing more to information security and privacy protection.

## 6. ACKNOWLEDGMENT

This work is supported by Research Foundation of China University of Petroleum-Beijing at Karamay （NO.XQZX20230030）and Talent Project of Tianchi Doctoral Program in Xinjiang Uygur Autonomous Region.